# Near-Infrared Image Dehazing Via Color Regularization

Chang-Hwan Son and Xiao-Ping Zhang

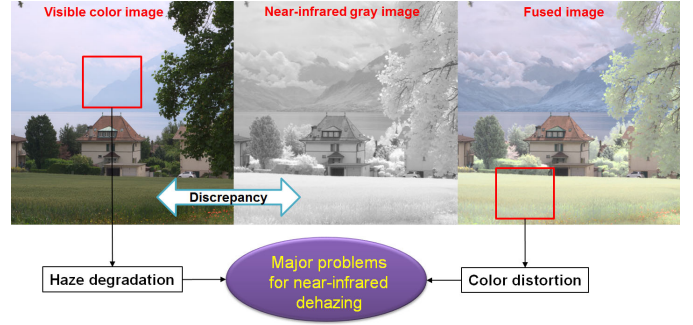

Fig. 1. An example of captured visible color image, near-infrared gray image, and fused image (upper part) and main issue for the near-infrared dehazing (lower part).

*Abstract*—Near-infrared imaging can capture haze-free near-infrared gray images and visible color images, according to physical scattering models, e.g., Rayleigh or Mie models. However, there exist serious discrepancies in brightness and image structures between the near-infrared gray images and the visible color images. The direct use of the near-infrared gray images brings about another color distortion problem in the dehazed images. Therefore, the color distortion should also be considered for near-infrared dehazing. To reflect this point, this paper presents an approach of adding a new color regularization to conventional dehazing framework. The proposed color regularization can model the color prior for unknown haze-free images from two captured images. Thus, natural-looking colors and fine details can be induced on the dehazed images. The experimental results show that the proposed color regularization model can help remove the color distortion and the haze at the same time. Also, the effectiveness of the proposed color regularization is verified by comparing with other conventional regularizations. It is also shown that the proposed color regularization can remove the edge artifacts which arise from the use of the conventional dark prior model.

## I. Introduction

According to Rayleigh physical scattering model [1], the intensity of the scattered light by the particles (e.g., dust, mist) in the atmosphere is inversely proportional to the wavelength of the incident light. Near-infrared lights have relatively long wavelengths between 700 nm and 1100 nm, compared to the visible lights that lie between 400nm and 700 nm. The more particles become denser, the visible lights get more scattered. These scattered lights called *airlight* [2] are blended with the reflected lights directly from the objects in the scene, thus degrading the visibility, contrast, or color fidelity of the captured visible color images. However, the near-infrared lights are less sensitive to the atmospheric scattering, in other words, near-infrared lights can reach the image sensors directly, and thus haze-free near-infrared images can be acquired.

Based on this scattering property, near-infrared imaging system, which captures the visible color images and the corresponding near-infrared gray images at the same time, was introduced in [3]. In this paper, two filters, that passes or blocks the near-infrared lights were used to capture both visible color and near-infrared images. Fig. 1 shows an example of the captured visible color image and near-infrared gray image. In the visible color image, we can notice that the contrast and details are almost lost in the haze region of the mountain and the sky, whereas the details can be captured in the near-infrared gray image, due to near-infrared lights scattering less. However, the captured near-infrared gray image looks unnatural, especially in the grass and the tree regions. This is because molecular structures cause the vegetation to 'glow' when viewed under the near-infrared lights. Also, the intensity of the near-infrared lights reflected by plants varies depending on the type of the plant; Please refer to [3] for more details.

### A. Main issue for near-infrared dehazing

As shown in Fig. 1, near-infrared imaging can provide a haze-free near-infrared gray image. It is tempting to use a naive fusion method that merely combines the chrominance planes from the visible color image with the near-infrared gray image in an opponent color space (e.g., $L^*a^*b^*$ [4] and decorrelated color space [5]). However, as already mentioned, there are discrepancies in the brightness and image structures between the visible color and near-infrared gray images, resulting in unnatural-looking colors, as shown in the rightmost image of Fig. 1. More sophisticated fusion methods based on multiresolution representation [6, 7] and principal component analysis [3] can be adopted. However, they all have color distortion problem. Another approach is using single image dehazing methods [8,9,10]. Given only visible color images, some constraints such as dark prior [8] and total variation [9,10]

C.-H. Son is with the Department of Electrical and Computer Engineering, Ryerson University, ON M5B2K3 Canada (E-mail: changhwan76.son@gmail.com).

X.-P. Zhang is with the Department of Electrical and Computer Engineering, Ryerson University, ON M5B2K3 Canada (E-mail: xzhang@ryerson.ca).

can help removing the haze. However, the dehazed image quality is not satisfactory. In some cases, haze still exists, or dehazed colors are distorted or unnatural. Therefore, the main issue for the near infrared dehazing is to remove the color distortion and the haze at the same time.

*B. Our approach*

To solve the main issue, this paper proposes a new color regularization to be incorporated into conventional dehazing framework. The concept of the proposed color regularization is so simple. For the proposed color regularization model, we first generate the unknown colors of the captured near infrared gray image. After that, we add a new color regularization term to the conventional dehazing framework to provide the color prior for unknown haze free image. However, some attentions are required during the color generation of the near-infrared gray image. That is, not only the created near-infrared color image should be natural-looking without color distortion, but it should also preserve the fine details of the captured near-infrared gray image. The newly added regularization term can provide the color information (i.e., color prior) for unknown haze-free images, so we call it *color regularization*.

The main focus of this paper is the color regularization. However, a new depth regularization is additionally introduced in this paper to propagate the colors and details induced from the use of the proposed color regularization into the depth map. Different from the conventional methods [9,10] where smoothness constraint between pixels in the depth map is used to model the depth regularization, the depth difference constraint between the consecutively estimated depth maps is adopted in this paper.

*C. Our contributions*

- First, our goal is to show how the proposed color regularization is useful for near-infrared dehazing. To be more concrete, we will show how effectively the proposed color regularization can remove the color distortion and the haze at the same time, which is the main issue for near-infrared dehazing, as shown in Fig. 1.
- Second, we show that the visual effects obtained from the near-infrared dehazing are not identical as those from the near-infrared coloring [3]. The proposed color regularization term is combined with the conventional dehazing framework, and thus visual color appearances of the dehazed visible images can be different from those of the colored near-infrared images. In other words, the dehazed visible color images are more vivid than the colored near-infrared images. In addition, the dehazed visible images have better gradation and color descriptions for the shadow and sky regions. It indicates that the near-infrared dehazing should be differentiated from the near-infrared coloring.
- Third, we show that the proposed color regularization can also remove the edge artifacts accompanied by the conventional dark prior model in [8]. The alpha matting algorithm is used in [11] to remove artifacts but it requires high computation complexity to create a Laplacian matrix [11,10]. If the width and height of input image is $W$ and $H$, the Laplacian matrix will be $WH \times WH$ in size. The proposed color regularization can avoid using the alpha matting algorithm.

## II. NOTATION

In this paper, bold lowercase is used to indicate column vectors. For example, $\mathbf{x}^v$ and $\mathbf{x}^{nir}$ indicate the column vectors that contain the pixel values of the captured visible color image and near-infrared gray image, respectively. The superscripts are used to differentiate between the two images. If the captured images are not grayscale, i.e., there are color channels, the superscript $c$ is additionally used to indicate the color channel like this $\mathbf{x}^{v(c)}$. Here, the superscript $c$ can be one of R, G, or B channels and $L^*$, $a^*$, or $b^*$ channels [4]. Other color channels (e.g., decorrelated color channels [5]) can be considered. To indicate vector elements (i.e., pixel values), the subscript $i$ is used like this $\mathbf{x}^v_i$.

## III. HAZE DEGRADATION MODEL

Before introducing our dehazing model, we first look at the haze degradation model [8], which can help us to understand how the haze in the images is formed. The haze degradation model is given by

$$\mathbf{x}^v_i = t_i \mathbf{x}^s_i + (1-t_i)\mathbf{x}^a \quad (1)$$

where the subscript $i$ denotes the pixel index. $\mathbf{x}^v_i$, $\mathbf{x}^s_i$, and $\mathbf{x}^a$ indicate column vectors with the size of $3 \times 1$. $\mathbf{x}^v_i$ and $\mathbf{x}^s_i$ contain $i$th R, G, and B pixel values of the captured haze image and unknown haze-free image (i.e., scene radiance), respectively; $\mathbf{x}^a$ is the atmospheric color light. Note that $\mathbf{x}^a$ is always identical irrespective of the pixel location ($i$); $t_i$ is a scalar value indicating the transmission, which describes the portion of the light that is not scattered and reaches the camera. Therefore, the equation (1) shows that the $i$th pixel values of the haze image $\mathbf{x}^v_i$ are given by blending the $i$th pixel values of the haze-free image $\mathbf{x}^s_i$ and the atmospheric color light $\mathbf{x}^a$, according to the transmission $t_i$. When the atmosphere is homogenous, the transmission $t_i$ can be expressed by $t_i = e^{-\eta d_i}$, where $\eta$ is the medium extinction coefficient and $d$ is the depth of scene. This means that the pixel values of the haze-free image $\mathbf{x}^s_i$ are attenuated exponentially with the scene depth $d_i$.

For simplicity, (1) can be rewritten [9], as follows.

$$\mathbf{x}^v_i - \mathbf{x}^a = t_i(\mathbf{x}^s_i - \mathbf{x}^a) \quad (2)$$

$$\ln(\mathbf{x}^v_i - \mathbf{x}^a) = \ln t_i + \ln(\mathbf{x}^s_i - \mathbf{x}^a) \quad (3)$$

where $\ln$ indicates the natural logarithm and it is used to avoid product term. Eq. (3) is further simplified as

$$\mathbf{u}_i^s = \mathbf{u}_i^d + \mathbf{u}_i^v \qquad (4)$$

where $\mathbf{u}_i^s$ and $\mathbf{u}_i^v$ are defined as $\ln(\mathbf{x}_i^s - \mathbf{x}^a)$ and $\ln(\mathbf{x}_i^v - \mathbf{x}^a)$, respectively, and $\mathbf{u}_i^d$ is expressed as $[u_i^d u_i^d u_i^d]^T$. Here, $T$ is transpose operator and $u_i^d$ is equal to $-\ln t_i$ when $\eta = 1$. Now, $\mathbf{x}^a$ is incorporated into $\mathbf{u}_i$ and $\mathbf{v}_i$, and thus the pixel index $i$ can be omitted, as follows.

$$\mathbf{u}^{s(c)} = \mathbf{u}^{d(c)} + \mathbf{u}^{v(c)} \qquad (5)$$

where $c$ indicates the color channel (e.g., R, G, and B channels). Note that $\mathbf{u}^{s(c)}$, $\mathbf{u}^{v(c)}$, and $\mathbf{u}^{d(c)}$ are the column vectors with the size of $N \times 1$. Here $N$ is the image size. Above equation shows that the haze degradation model can be represented by the image-based operation, which is different from (1) based on the element-wise operation.

IV. OUR PREVIOUS STUDY ON NEAR-INFRARED COLORING

As briefly mentioned in the Introduction, the proposed color regularization model requires the colored version of the near-infrared gray image. Now, we will outline our previous study on the near-infrared coloring [12]. This near-infrared coloring method used a contrast-preserving mapping model to solve the discrepancy problem. Natural-looking colors, fine details, and high contrast can be rendered on the created near-infrared color images. Moreover, the color distortion, as shown in Fig. 1, can be avoided. For this reason, the near-infrared coloring method in [12] will be used to model the proposed color regularization.

The proposed near-infrared coloring method is roughly divided by two steps: the contrast-preserving mapping and color transfer. First, the contrast-preserving mapping finds the relation between the luminance planes for the visible color image and near-infrared gray image. And then, it generates a new, artificially generated near-infrared gray image, which is not the same as the captured near-infrared gray image. Second, the color transfer method corrects the colors of the visible color image using the mapping relation, and then adds the corrected colors to the newly created near-infrared gray image, not the captured near-infrared gray image.

*A. Contrast-preserving mapping*

To derive a mapping relation between the two luminance planes for the visible color image and near-infrared gray image, the following formula can be used.

$$\min_{\boldsymbol{\alpha}_i} \left\| \mathbf{W}_i^{1/2}(\mathbf{p}_i - [\mathbf{q}_i \; \mathbf{1}]\boldsymbol{\alpha}_i) \right\|_2^2 + \mu_c \left\| \boldsymbol{\alpha}_i - \boldsymbol{\alpha}_i^0 \right\|_2^2 \qquad (6)$$

where $\|\cdot\|_p$ indicates the $p$-norm. $\mathbf{p}_i$ and $\mathbf{q}_i$ are the column vectors that contain the pixel values of the extracted patches from the luminance planes of the visible color and near-infrared images at the $i$th pixel location, respectively. $\mathbf{1}$ is the column vector filled with one and $\mathbf{W}_i$ is a diagonal matrix consisting of weights that are inversely proportional to the distance between the center pixel location $i$ and its neighboring pixel location. If the extracted patch has a size $m \times m$ (e.g., $5 \times 5$), the dimensions of the $\mathbf{p}_i$ and $\mathbf{W}_i$ will be $m^2 \times 1$ and $m^2 \times m^2$, respectively. Vector $\boldsymbol{\alpha}_i^T = [\alpha_{i,1} \; \alpha_{i,2}]$ to be estimated here contains two vector elements indicating slope and bias, respectively. Therefore, the data fidelity term $\left\| \mathbf{W}_i^{1/2}(\mathbf{p}_i - [\mathbf{q}_i \; \mathbf{1}]\boldsymbol{\alpha}_i) \right\|_2^2$ can be regarded as a linear mapping. In other words, the near-infrared luminance patch $\mathbf{q}_i$ is mapped to the visible luminance patch $\mathbf{p}_i$ without any constraints. Adding a local contrast-preserving regularization term $\left\| \boldsymbol{\alpha}_i - \boldsymbol{\alpha}_i^0 \right\|_2^2$ prevents this. $\boldsymbol{\alpha}_i^0$ is given from the extracted near-infrared and visible-luminance patches, according to a local contrast measure [12,13].

Given the estimated $\boldsymbol{\alpha}_i$, a new, artificially created near-infrared luminance image can be obtained, as follows:

$$\mathbf{x}_i^{o(L^*)} = \mathbf{x}_i^{nir} \alpha_{i,1} + \alpha_{i,2} \qquad (7)$$

where $\mathbf{x}_i^{nir}$ and $\mathbf{x}_i^{o(L^*)}$ are the captured near-infrared luminance image and the newly created near-infrared luminance image, respectively. Here, $\alpha_i$ is the linear-mapping relation between the luminance images.

*B. Color transfer*

The colors of the newly created near-infrared luminance image can be derived from the colors (i.e., the chrominance planes) of the visible color image, according to the mapping relation $\boldsymbol{\alpha}_i$, as follows,

$$\mathbf{x}_i^{o(a^*)} = \mathbf{x}_i^{v(a^*)} / \alpha_{i,1} \quad \text{and} \quad \mathbf{x}_i^{o(b^*)} = \mathbf{x}_i^{v(b^*)} / \alpha_{i,1} \qquad (8)$$

where $\mathbf{x}_i^{v(a^*)}$ and $\mathbf{x}_i^{v(b^*)}$ indicate the two chrominance planes of the visible color image and $\mathbf{x}_i^{o(a^*)}$ and $\mathbf{x}_i^{o(b^*)}$ are the two chrominance planes to be added to the newly created near-infrared luminance image $\mathbf{x}_i^{o(L^*)}$. Thus, the above equation tells us that the unknown chrominance planes $\mathbf{x}_i^{o(a^*)}$ and $\mathbf{x}_i^{o(b^*)}$ for the newly created infrared luminance image $\mathbf{x}_i^{o(L^*)}$ can be obtained by dividing the chrominance planes for the visible color image by $\alpha_{i,1}$, the mapping relation. Equation (8) is derived from the contrast-preserving linear mapping—i.e., by $\mathbf{x}_i^{nir} \alpha_{i,1} \approx \mathbf{x}_i^{o(L^*)}$, which reveals that the unknown chrominance planes for the newly created near-infrared gray image can be defined as the contrast-enhanced version of the chrominance planes for the visible color image. The final colored near-infrared image $\mathbf{x}^o$ can be obtained by combining the newly created near-infrared gray image $\mathbf{x}_i^{o(L^*)}$ with the chrominance planes $\mathbf{x}_i^{o(a^*)}$ and $\mathbf{x}_i^{o(b^*)}$, and then transformed into the RGB color space via opponent color space conversion. In this paper, the decorrelate color space [5] was used.

## V. Proposed Near-Infrared Dehazing

Assuming that the unknown colors of the captured near-infrared gray image are defined, according to (6)-(8), the proposed near-infrared dehazing can be modeled, as follows:

$$\min_{\mathbf{u}_{t+1}^{s(c)},\mathbf{u}_{t+1}^{d(c)}} \lambda_1 \left\| \mathbf{u}_{t+1}^{s(c)} - (\mathbf{u}^{v(c)} + \mathbf{u}_{t+1}^{d(c)}) \right\|_2^2 + \underbrace{\lambda_2 \left\| \mathbf{u}_{t+1}^{s(c)} - \mathbf{u}^{o(c)} \right\|_2^2}_{\text{Color Regularization}} + \sum_{j=1}^{2} \left\| \mathbf{f}^j \oplus \mathbf{u}_{t+1}^{s(c)} \right\|_1 + \underbrace{\lambda_3 \left\| \mathbf{u}_{t+1}^{d(c)} - \mathbf{u}_t^{d(c)} \right\|_2^2}_{\text{Depth Regularization}} \quad (9)$$

where the vectors $\mathbf{u}^{s(c)}$, $\mathbf{u}^{v(c)}$, and $\mathbf{u}^{d(c)}$ are the same as the ones defined in (5), but the two vectors $\mathbf{u}^{s(c)}$ and $\mathbf{u}^{d(c)}$ are iteratively updated with an iteration number $t$. The vector $\mathbf{u}^{o(c)}$ includes the information about the colored version of the near-infrared gray image, thus it is not updated through the iteration. According to the haze degradation model, as shown in (3) and (4), $\mathbf{u}_i^{o(c)}$ is given by $\ln(\mathbf{x}_i^{o(c)} - \mathbf{x}^{a(c)})$ where $\mathbf{x}^o$ is the colored version of the captured near-infrared gray image according to (6)-(8), and the atmospheric color light $\mathbf{x}^a$ is given by the dark channel prior [8], which will be described later. The symbol $\oplus$ denotes convolution operator and $\mathbf{f}^j$ indicates the horizontal or vertical gradient filters. $\lambda_1$, $\lambda_2$, and $\lambda_3$ are penalty parameters; $c$ indicates one of the R, G, or B channels.

In (9), the first term indicates the haze degradation model defined in (5). The model imposes a constraint that the captured haze image should be modeled by the blending of the unknown haze-free image and atmospheric color light. The second term is the proposed color regularization. As mentioned in the Introduction, there are serious discrepancies in brightness and image structures between the visible color image and near-infrared gray image. This discrepancy generates another color distortion problem during near-infrared dehazing. Our color regularization term prevents the colors of the unknown haze-free image from largely deviating from the rendered colors of the captured near-infrared gray image. The third term is the gradient regularization. It is well-known that the gradient distribution of haze-free images can be modeled by Laplacian or hyper-Laplacian probability distributions in a logarithm domain [14]. The total variation norm in third term can reflect the statistical gradient distribution of haze-free images. The last term is another proposed depth regularization which ensures that the currently estimated depth map should not be largely deviated from the previously estimated depth map. Different from the conventional methods [9,10] where smoothness constraint between neighborhood pixels in the depth map is used to model the depth regularization, the proposed dehazing model adopts the depth difference constraint between the consecutively estimated depth maps. This depth regularization term can propagate the details and colors of the estimated haze-free images into the updated depth maps during iteration. The role of this depth regularization will be discussed in section V.C.2.

### A. Advantages of the proposed color regularization

- If $\lambda_2$ is set with zero, (9) becomes the single image dehazing model. In this case, only available information is the captured haze image. The prior information is the gradient distribution, as shown in the third term. Thus, it is difficult to estimate the unknown haze-free image, atmospheric color light, and depth map at the same time. Even though dark channel prior can be used to initialize depth map and atmospheric color light, in most cases, it fails to provide satisfactory results. However, the use of the proposed color regularization can provide good initial point for the unknown haze-free image, thereby leading to reach a good solution.

- Instead of the proposed color regularization, other regularizations can be considered. For image-pair-based restoration, gradient difference regularization has been widely used [15,16]. Recently, this regularization was also applied for near-infrared dehazing [17]. The gradient difference regularization is represented as $\sum_{j=1}^{2} \left\| \mathbf{f}^j \oplus \mathbf{u}_{t+1}^{s(c)} - \mathbf{f}^j \oplus \mathbf{u}^{nir} \right\|_1$ where $\mathbf{u}^{nir}$ is given by $\ln(\mathbf{x}_i^{nir} - \mathbf{x}^a)$, according to the haze degradation model defined as in (3) and (4). This gradient difference regularization can be replaced by the proposed color regularization. However, note that the captured near-infrared image $\mathbf{x}^{nir}$ is grayscale, and thus color information cannot be provided. Only gradient information is available. In contrast, the proposed color regularization can provide color and gradient information, which leads to improvement in visual color appearance. In addition, the edge artifacts that appear on the initial haze-free image can be reduced with the proposed color regularization. However, the gradient difference regularization cannot remove the edge artifacts.

### B. Near-infrared dehazing vs. near-infrared coloring

The proposed color regularization additionally uses the colored version of the near-infrared gray image. At this time, we can have a question whether there is a difference between the near-infrared dehazing and near-infrared coloring [3,12]. As shown in (9), the proposed dehazing model tries to satisfy both haze degradation model (i.e., first term) and color regularization (i.e., second term). On the other hand, near-infrared coloring [3,12] excludes the haze degradation model. Therefore, the estimated haze-free image will be different from the colored version of the near-infrared gray image. Specifically, the estimated haze-free images can be more colorful than the colored near-infrared images. Moreover, the estimated haze-free images can have better gradation and color descriptions than the colored near-infrared images, especially for the shadow and sky regions, which will be checked in the experimental results.

*C. Implementation*

The equation (9) can be divided into two minimization problems, according to alternating minimization scheme [18], as follows:

$$\min_{\mathbf{u}_{t+1}^{s(c)}} \lambda_1 \left\| \mathbf{u}_{t+1}^{s(c)} - (\mathbf{u}^{v(c)} + \mathbf{u}_t^{d(c)}) \right\|_2^2 + \lambda_2 \left\| \mathbf{u}_{t+1}^{s(c)} - \mathbf{u}^{o(c)} \right\|_2^2 + \sum_{j=1}^{2} \left\| \mathbf{f}^j \oplus \mathbf{u}_{t+1}^{s(c)} \right\|_1 \quad (10)$$

$$\min_{\mathbf{u}_{t+1}^{d(c)}} \left\| \mathbf{u}_{t+1}^{d(c)} - (\mathbf{u}_{t+1}^{s(c)} - \mathbf{u}^{v(c)}) \right\|_2^2 + \lambda_3 \left\| \mathbf{u}_{t+1}^{d(c)} - \mathbf{u}_t^{d(c)} \right\|_2^2 \quad (11)$$

Given the initial depth map $\mathbf{u}_{t=0}^{d(c)}$, the unknown haze-free image $\mathbf{u}_{t+1}^{s(c)}$ is first estimated by minimizing (10), and then by inserting the estimated haze free image $\mathbf{u}_{t+1}^{s(c)}$ into (11) to update the initial depth map $\mathbf{u}_{t=0}^{d(c)}$. Then, the updated depth map $\mathbf{u}_{t+1}^{d(c)}$ is inserted into (10) to estimate the unknown haze-free image again. This process is iteratively repeated until a stopping condition is satisfied. In this paper, maximum iteration number is used for the stopping condition.

*1) Initialization:* To solve (10) and (11), we first need initialize three vectors: depth map $\mathbf{u}_{t=0}^{d(c)}$, atmospheric color light $\mathbf{x}^a$, and colored near-infrared image $\mathbf{u}^{o(c)}$. First, to provide the $\mathbf{u}_{t=0}^{d(c)}$ and $\mathbf{x}^a$, dark channel prior [8] was utilized. The dark channel prior is a kind of statistics on the haze-free outdoor images. In most non-sky patches of the haze-free outdoor images, at least one color channel has very low intensity value. Based on this observation, the dark channel is defined as

$$\mathbf{x}_i^{dark} = \min_c \min_{j \in \Omega_i} \mathbf{x}_j^{s(c)} \approx 0 \quad (12)$$

where $j$ indicates the neighborhood pixel index that belongs to a local patch $\Omega$ centered at pixel index $i$. The above equation indicates that the intensity values of the dark channel vector $\mathbf{x}_i^{dark}$ are distributed around zero. Based on this dark channel prior, the transmission $t_i$ can be derived, as follows

$$t_i = 1 - \min_c \min_{j \in \Omega_i} \left( \frac{\mathbf{x}_j^{v(c)}}{\mathbf{x}^a} \right) \quad (13)$$

where the atmospheric color light $\mathbf{x}^a$ is given by picking and averaging the top 0.1% brightest pixels in $\mathbf{x}_i^{dark}$; Please refer to [8] for more details about the transmission and the atmospheric color light. Given the transmission $t_i$, the initial depth map $\mathbf{u}_{t=0}^{d(c)}$ can be calculated based on the relation between the transmission and scene depth, i.e., $\mathbf{u}_{i,t=0}^{d(c)} = -\ln t_i$. However, as discussed in [8,9], it is assumed that the transmission $t_i$ is constant in a local patch, and thus this leads to the edge artifacts on the initial dehazed images. To remove this edge artifacts, alpha matting algorithm was used in [8,11], however it requires high computation complex.

Second, the colored near-infrared image $\mathbf{x}^{o(c)}$ is provided, according to (6)-(8). To obtain the $\mathbf{u}^{o(c)}$, $\mathbf{x}^{o(c)}$ is first subtracted from the atmospheric color light $\mathbf{x}^a$, and then converted into the logarithmic domain, according to the haze degradation model, as follows,

$$\mathbf{u}_i^{o(c)} = \ln(\mathbf{x}_i^{o(c)} - \mathbf{x}^{a(c)}) \quad (14)$$

*2) Numerical solution:* The equation (10) is well-known as total variation problem. There are many optimization techniques to solve (10). In this paper, variable-splitting technique [18,19] is adopted. Certainly, other minimization techniques [20,21] could also be considered. The equation (10) can be divided into two minimization problems with the auxiliary vector $\mathbf{w}$, as follows,

$$\min_{\mathbf{u}_{t+1}^{s(c)}, \mathbf{w}} \frac{\lambda_0}{2} \left\{ w_1 \left\| \mathbf{u}_{t+1}^{s(c)} - (\mathbf{u}^{v(c)} + \mathbf{u}_t^{d(c)}) \right\|_2^2 + w_2 \left\| \mathbf{u}_{t+1}^{s(c)} - \mathbf{u}^{o(c)} \right\|_2^2 \right\} + \sum_{j=1}^{2} \left\| \mathbf{w}^j \right\|_1 + \frac{\beta}{2} \sum_{j=1}^{2} \left\| \mathbf{w}^j - \mathbf{f}^j \oplus \mathbf{u}_{t+1}^{s(c)} \right\|_2^2 \quad (15)$$

where $\lambda_0 w_1 / 2$ and $\lambda_0 w_2 / 2$ correspond to $\lambda_1$ and $\lambda_2$ shown in (10), respectively. $w_1$ and $w_2$ are weighting values and their sum should equal to one. The equation (15) has two unknown vectors $\mathbf{u}_{t+1}^{s(c)}$ and $\mathbf{w}$, which are iteratively updated via alternating minimization. In other words, given a fixed $\mathbf{w}$, $\mathbf{u}_{t+1}^{s(c)}$ is minimized with a close-form solution, and then $\mathbf{w}$ is minimized with a shrinkage operation for a fixed $\mathbf{u}_{t+1}^{s(c)}$. This processing continues until a stop criterion is satisfied. During the iteration, the penalty parameter $\beta$ should be increased for convergence. Please refer to [18,19] for more details.

The equation (11) has a closed-form solution, as follows,

$$\mathbf{u}_{t+1}^{d(c)} = \frac{1}{\lambda_3} \left\{ (\mathbf{u}_{t+1}^{s(c)} - \mathbf{u}^{v(c)}) + \lambda_3 \mathbf{u}_t^{d(c)} \right\} \quad (16)$$

The above equation shows that the updated depth map $\mathbf{u}_{t+1}^{d(c)}$ is weighted averaging of the previous depth map $\mathbf{u}_t^{d(c)}$ and the currently estimated depth map $(\mathbf{u}_{t+1}^{s(c)} - \mathbf{u}^{v(c)})$, which is defined by the haze degradation model. Note that the currently estimated depth map $\mathbf{u}_{t+1}^{d(c)}$ includes the updated haze-free image $\mathbf{u}_{t+1}^{s(c)}$. Through the depth map updating via (16), initial depth map $\mathbf{u}_{t=0}^{d(c)}$ can be improved because the updated $\mathbf{u}_{t+1}^{s(c)}$ includes the colored version of the near-infrared gray image $\mathbf{u}^{o(c)}$, which includes fine details and natural-looking colors. At the next iteration, the updated depth map naturally influences the haze-free image prediction, thereby upgrading the initial dehazed image $\mathbf{u}_{t=0}^{s(c)} = \mathbf{u}^{v(c)} - \mathbf{u}_{t=0}^{d(c)}$. In other words, the initial dehazed image $\mathbf{u}_{t=0}^{s(c)}$ was produced by applying the single image dehazing method [8], and thus its image quality is poor. However, the use of the proposed color regularization can transfer the fine details and natural-looking colors of the created near-infrared image to the initial dehazed image $\mathbf{u}_{t=0}^{s(c)}$, thereby producing high-quality dehazed images. The proposed color regularization provides the color prior for the unknown haze-free image and the proposed depth regularization propagates the details and natural-looking colors of the estimated haze-free images $\mathbf{u}_{t+1}^{s(c)}$ into the updated depth maps $\mathbf{u}_{t+1}^{d(c)}$. This is the key of our method.

Our near-infrared dehazing method is summarized in Algorithm I. Given the colored near-infrared image $\mathbf{x}^o$, atmospheric color light $\mathbf{x}^a$, and transmission $t_i$, the unknown vectors $\mathbf{u}_{t+1}^{s(c)}$ and $\mathbf{u}_{t+1}^{d(c)}$ are iteratively updated for each color channel $c$. After reaching the maximum iteration of the inner loop $t_{max}$, the dehazed image $\mathbf{x}^{s(c)}$ for the channel $c$ is obtained by manipulating $\mathbf{u}^{s(c)}$. When outer loop finishes, the final dehazed color image $\mathbf{x}^s$ is obtained. In Algorithm I, $c_{max}$ and $t_{max}$ are assigned by 3 and 7, respectively. The edge artifacts on initial dehazed images can be removed after $t_{max} = 7$. The parameters in (15) and (16) are set with $\lambda_0 = 10^5$, $w_1 = 0.8$, $w_2 = 0.2$, and $\lambda_3 = 1$.

---

**Algorithm I: Near-infrared dehazing method**

**Input:** Near-infrared gray image $\mathbf{x}^{nir}$ and visible color image $\mathbf{x}^v$
**Output:** Dehazed visible color image $\mathbf{x}^s$
**Initialization:**
- Generate the colored near-infrared image $\mathbf{x}^o$, according to (6)-(8)
- Find the transmission $t_i$ and atmospheric color light $\mathbf{x}^a$ using (12) and (13)
- Define $\mathbf{u}^{o(c)}$, $\mathbf{u}^{v(c)}$, and $\mathbf{u}_{t=0}^{d(c)}$ from $\mathbf{x}^o$, $\mathbf{x}^v$, and $t_i$, respectively, according to the haze degradation model, as shown in (3) and (4)
- Initialize the parameter $w_1$, $w_1$, $\lambda_0$, $\lambda_3$

**Near-infrared dehazing:**
  **for** $c = 1,2,..,c_{max}$
    **for** $t = 1,2,..,t_{max}$
- Update $\mathbf{u}_{t+1}^{s(c)}$ via (15)
- Update $\mathbf{u}_{t+1}^{d(c)}$, according to (16)

    **end**
- Calculate $\mathbf{x}^{s(c)} = \mathbf{x}^{a(c)} - e^{\eta \mathbf{u}^{s(c)}}$

  **end**
**Return** $\mathbf{x}^s$

---

## VI. EXPERIMENTAL RESULTS

In this paper, three types of experiments are conducted. First, visual effects, according to the use of the near-infrared dehazing and coloring, will be compared. This experiment will give us the difference between the near-infrared dehazing and coloring. Next, different regularizations will be tested using our dehazing model to confirm how the proposed color regularization is more effective than other regularizations for near-infrared dehazing. Finally, visual quality comparison for different dehazing methods will be followed.

### A. Near-infrared dehazing vs. coloring

First, we confirm that the colored near-infrared images via near-infrared coloring [12] have different visual appearances from the dehazed visible color images. As shown in Figs. 2-5, the colors of the dehazed visible images are vividly rendered, however the colored near-infrared images are slightly diluted. In addition, the cloud, sky, and shadow descriptions of the dehazed visible color images are better. To check these, see the red boxes. These results arise from the difference between the near-infrared coloring and near-infrared dehazing. The purpose of the near-infrared coloring is to add colors to the captured near-infrared gray images, and thus the visual appearances of the colored near-infrared images are mainly influenced by the overall brightness of the captured near-infrared gray images. In contrast, the purpose of the near-infrared dehazing is to remove the haze from the captured visible color images, based on the haze degradation model. Even though the proposed color regularization additionally utilizes the colored near-infrared images, the value of $w_1$ is largely set than that of $w_2$, which means that the visual color appearances of the dehazed images are dominantly influenced by the captured visible color images.

There are mainly two approaches of removing the haze. One is to use the near-infrared coloring, which adds colors to the haze-free near-infrared gray images. The other is to use the near-infrared dehazing, which removes the haze from the visible color images. The near-infrared coloring method can be a good solution to remove the haze, as shown in Figs. 2-5. The main difference between the near-infrared coloring and dehazing depends on how to render the visual color appearance. Whether the haze degradation model is used can make a difference in rendered visual color appearances.

### B. Visual effects according to the used regularizations

Now let us look at how the used different regularizations can influence the visual quality. Fig. 6 shows the dehazed images using different regularizations: dark channel prior [8], gradient regularization [14], and gradient difference regularization [17]. In Fig. 6, the first row shows the initial dehazed images using the single image dehazing method where the dark channel prior was used. However alpha matting algorithm [11] was not applied to check pure visual effects, according to the used regularizations. As shown in the figures, the visual qualities of the dehazed images are not satisfactory. The overall brightness of the dehazed images is dark and the edge artifacts arise, as shown in the yellow boxes. The reasons are already discussed in [8]. The second row shows the dehazed images using gradient regularization. In other words, $\lambda_2$ is set with zero in (9), which indicates that the proposed dehazing model becomes equal to the single image dehazing model. In this case, the gradient regularization, in other words, only the third term is considered. The use of the gradient regularization makes the initial dehazed images smooth, and the strength of the edge artifacts can be reduced. However, overall sharpness can be decreased as well. The original purpose of using the gradient regularization is to suppress the noise amplified after image dehazing [9,10]. The last row shows the dehazed images using gradient difference regularization. In (9), the proposed color regularization term was replaced by the gradient different regularization, which is expressed as $\sum_{j=1}^{2}\left\|\mathbf{f}^j \oplus \mathbf{u}_{t+1}^{s(c)} - \mathbf{f}^j \oplus \mathbf{u}^{nir}\right\|_1$. This regularization term can force the gradients of the initial dehazed images to be close to those of the captured near-infrared gray images. Thus, the edge strength of the initial dehazed images can be increased, as shown in the red boxes. However, the overall color appearances cannot be changed significantly. In other words, both gradient and gradient difference regularizations are highly dependent on the visual quality levels of the initial dehazed images. On the other hand, the proposed color regularization can provide the colored versions of the near-infrared gray images to initialize unknown haze-free images. This can result in natural-looking colors and fine details. This is why the proposed color regularization can provide better results, as shown in Figs. 2(c), 3(c), 4(c), and 5(c). Especially, the edge artifacts can be removed without using the alpha matting algorithm [11], which requires high computation complexity [9]. In addition, the color and details can be improved significantly.

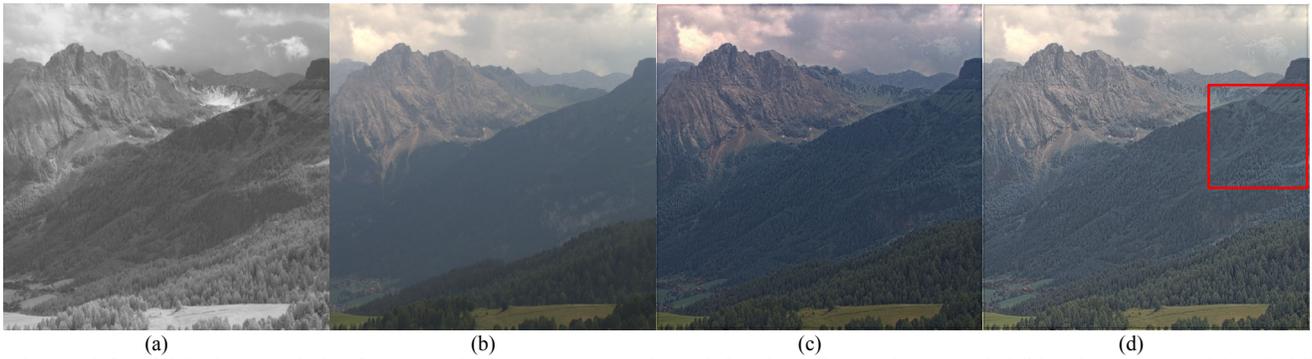

Figure 2. Near-infrared dehazing vs. coloring for 'Mountain' image; (a) captured near-infrared gray image, (b) captured visible color image, (c) dehazed visible image with the proposed method, and (d) colored near-infrared gray image [12] (Please zoom in to the images to check details and colors).

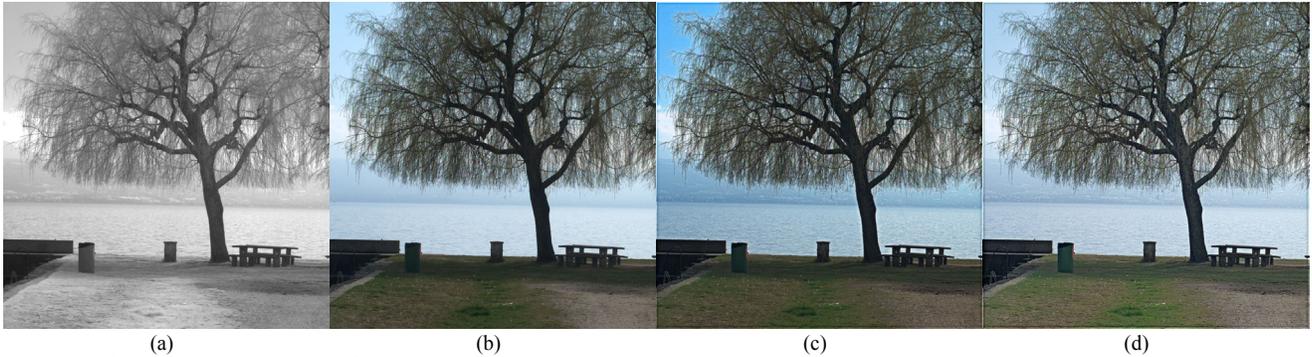

Figure 3. Near-infrared dehazing vs. coloring for 'Lake' image; (a) captured near-infrared gray image, (b) captured visible color image, (c) dehazed visible image with the proposed method, and (d) colored near-infrared gray image [12] (Please zoom in to the images to check details and colors).

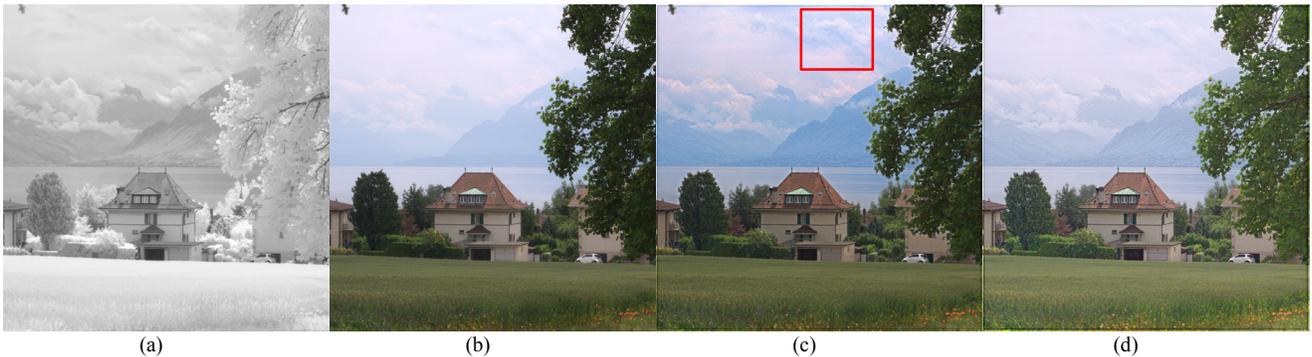

Figure 4. Near-infrared dehazing vs. coloring for 'House' image; (a) captured near-infrared gray image, (b) captured visible color image, (c) dehazed visible image with the proposed method, and (d) colored near-infrared gray image [12] (Please zoom in to the images to check details and colors).

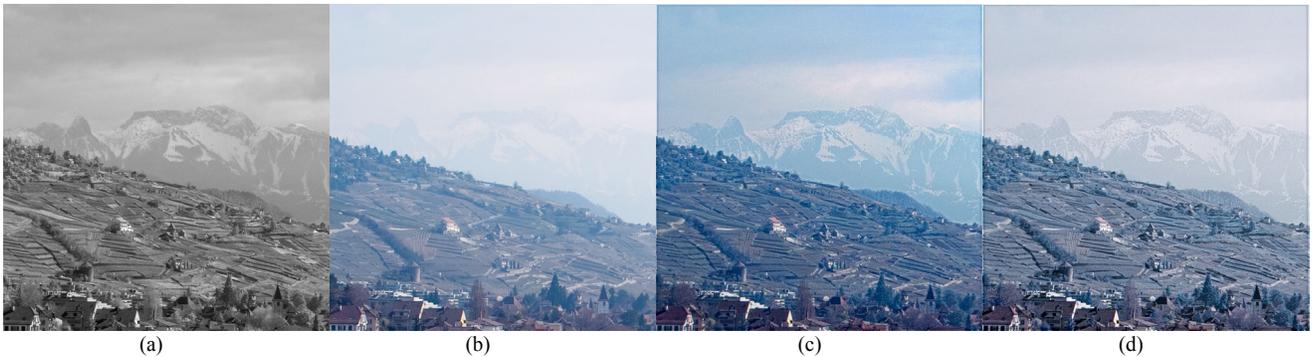

Figure 5. Near-infrared dehazing vs. coloring for 'Village' image; (a) captured near-infrared gray image, (b) captured visible color image, (c) dehazed visible image with the proposed method, and (d) colored near-infrared gray image [12] (Please zoom in to the images to check details and colors).

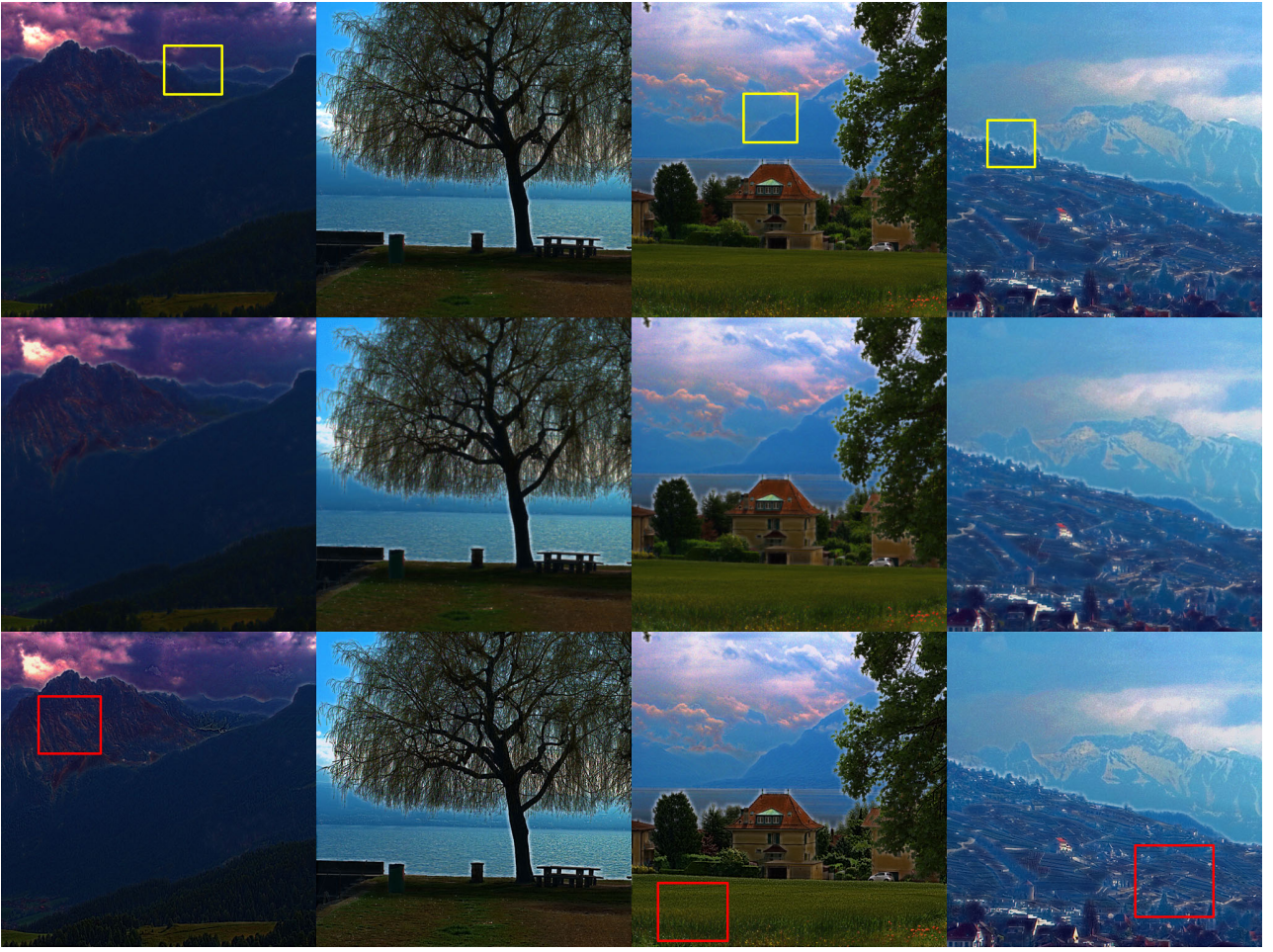

Figure 6. Experimental results for the used regularizations; (a) initial dehazed images via dark channel prior where alpha matting was not applied (first row), (b) dehazed images using gradient regularization (second row), and (c) dehazed images using gradient difference regularization (last row).

### C. Visual quality comparison

Figs. 7-10 show the dehazed images with the conventional single image dehazing method [8], image-pair-based dehazing methods [6,17], and the proposed method. The dehazed images of the proposed method are the same as the ones in Figs. 2-5. The single image dehazing method [8] using the dark channel prior is the same as the one used in the previous section, however the alpha matting algorithm [11] was applied to refine the initial dehazed images in the first row of Fig. 6. The use of the alpha matting algorithm can improve the visual quality of the dehazed images, as shown in Figs. 7(a), 8(a), 9(a), and 10(a). Especially, the edge artifacts can be significantly reduced. However, even though the single image dehazing method is one of the state-of-the-art methods, haze cannot be completely removed, as shown in the red boxes of Figs. 9(a) and 10(a). Most of all, the used alpha matting algorithm requires high complex computation to generate a Laplacian matrix [11]. If the width and height of input image is $W$ and $H$, the Laplacian matrix will be $WH \times WH$ in size. Figs. 7(b), 8(b), 9(b), and 10(b) show the dehazed images using image-pair-based dehazing method [6] where the near-infrared gray images and visible color images are both utilized. In this method, the details of the near-infrared gray images are

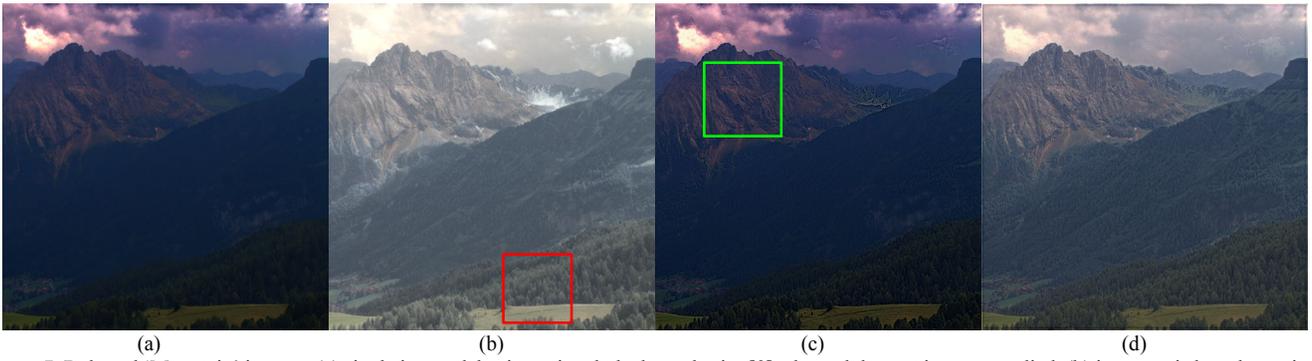

(a) (b) (c) (d)

Figure 7. Dehazed 'Mountain' images; (a) single image dehazing using dark channel prior [8] where alpha matting was applied, (b) image-pair-based near-infrared dehazing using multiresolution representation [6], and (c) image-pair-based near-infrared dehazing using gradient difference regularization [17], and (d) proposed method (Please zoom in the images to check details and colors).

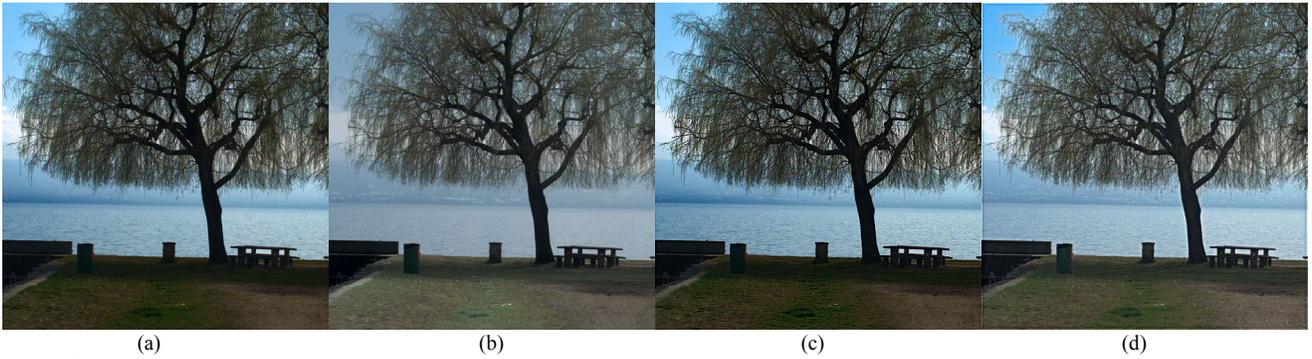

(a) (b) (c) (d)

Figure 8. Dehazed 'Lake' images; (a) single image dehazing using dark channel prior [8] where alpha matting was applied, (b) image-pair-based near-infrared dehazing using multiresolution representation [6], and (c) image-pair-based near-infrared dehazing using gradient difference regularization [17], and (d) proposed method (Please zoom in the images to check details and colors).

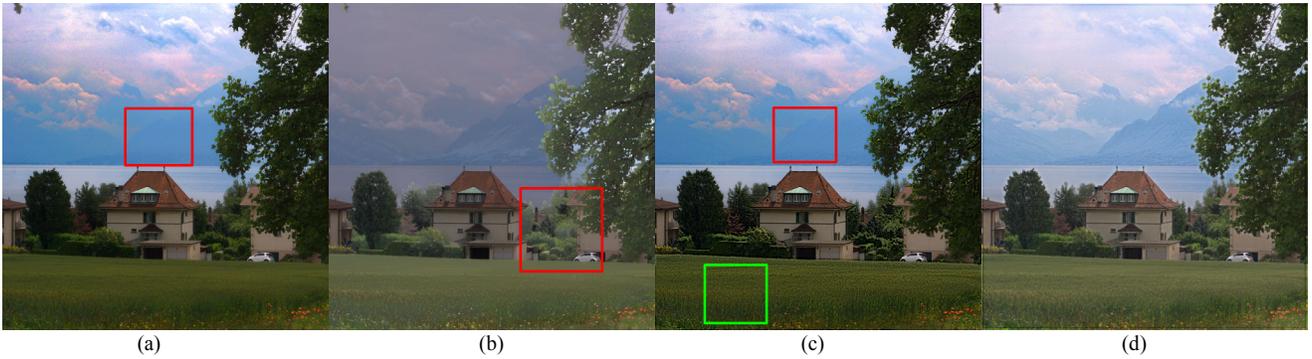

(a) (b) (c) (d)

Figure 9. Dehazed 'House' images; (a) single image dehazing using dark channel prior [8] where alpha matting was applied, (b) image-pair-based near-infrared dehazing using multiresolution representation [6], and (c) image-pair-based near-infrared dehazing using gradient difference regularization [17], and (d) proposed method (Please zoom in the images to check details and colors).

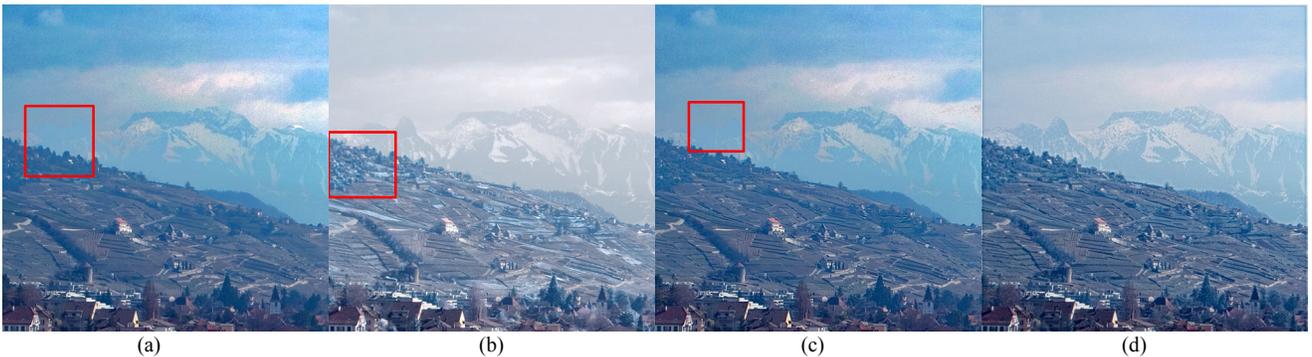

(a) (b) (c) (d)

Figure 10. Dehazed 'Village' images; (a) single image dehazing using dark channel prior [8] where alpha matting was applied, (b) image-pair-based near-infrared dehazing using multiresolution representation [6], and (c) image-pair-based near-infrared dehazing using gradient difference regularization [17], and (d) proposed method (Please zoom in the images to check details and colors).

TABLE I. DEFINITION OF THE THREE MEASURES

- $ISS = (\sigma_{\mathbf{x}^s\mathbf{x}^r} + c_3)/(\sigma_{\mathbf{x}^s}\sigma_{\mathbf{x}^r} + c_3)$ where $\sigma_{\mathbf{x}^s}$, $\sigma_{\mathbf{x}^r}$, and $\sigma_{\mathbf{x}^s\mathbf{x}^r}$ indicate the standard deviations and correlation coefficient for dehazed and reference images $\mathbf{X}^s$ and $\mathbf{X}^r$, respectively. $c_3$ is a constant value to avoid numerical instability [22].
- The ISS value ranges between 0 and 1. Here, '1' indicates the perfect match in the image structures between input images, whereas '0' indicates perfect mismatch.
- $CD = \sum_{i=1}^{N}\sqrt{\sum_{c=1}^{3}(\mathbf{x}_i^{s(c)} - \mathbf{x}_i^{r(c)})^2} / N$ where $\mathbf{X}^s$ and $\mathbf{X}^r$ indicates the dehazed and reference color images, respectively. The color channel $c$ indicates one of the $L^*, a^*,$ or $b^*$ color spaces [4].
- The CD value is greater than or equal to zero. When the colors of input images are identical, the CD value will be zero and it will increase as the colors become more different.
- $CF = \sigma_{ab} + 0.94u_C$, where $\sigma_{ab}$ and $u_C$ are related to the standard deviation of the chrominance planes and mean value of the chroma image, respectively [23].
- The CF will have a small value when an input test image looks grayish. Note that reference CF values are unknown. Thus, we cannot say that higher CF values indicate better visual quality. However, the CD measure includes the chroma difference, and thus it is expected that the reference CF values correspond to the smallest CD values.

TABLE II. QUANTITATIVE EVALUATION

| Method | Measures | Mountain | Lake | House | Village | AVG |
|---|---|---|---|---|---|---|
| Dark Channel Prior [8] | ISS | 0.8619 | 0.7306 | 0.8331 | 0.9106 | **0.8341** |
| | CD | 18.3315 | 10.2772 | 10.1315 | 15.3934 | 13.5334 |
| | CF | 15.2453 | 21.6904 | 26.3176 | 40.3445 | 25.8995 |
| Multiresolution Representation [6] | ISS | 0.9482 | 0.9447 | 0.9760 | 0.9738 | 0.9607 |
| | CD | 13.3806 | 5.4886 | 9.3962 | 6.0070 | 8.5681 |
| | CF | 11.2660 | 17.6496 | 20.2273 | 30.1572 | **19.8250** |
| Gradient Difference Regularization [17] | ISS | 0.9066 | 0.8521 | 0.8719 | 0.8538 | 0.8711 |
| | CD | 18.5545 | 11.0302 | 10.6415 | 15.9251 | **14.0378** |
| | CF | 15.1726 | 21.5641 | 26.0573 | 40.6340 | 25.8570 |
| Near-Infrared Coloring [12] | ISS | 0.9454 | 0.9308 | 0.9496 | 0.9700 | 0.9489 |
| | CD | 7.9292 | 7.5664 | 12.135 | 9.6295 | 9.3150 |
| | CF | 13.5218 | 19.8453 | 28.0196 | 26.0706 | 21.8643 |
| Proposed Method | ISS | 0.9388 | 0.9037 | 0.9296 | 0.9567 | 0.9322 |
| | CD | 4.2215 | 3.5201 | 4.6333 | 14.7750 | 6.7875 |
| | CF | 12.9675 | 19.2915 | 23.9960 | 36.8777 | 23.2832 |

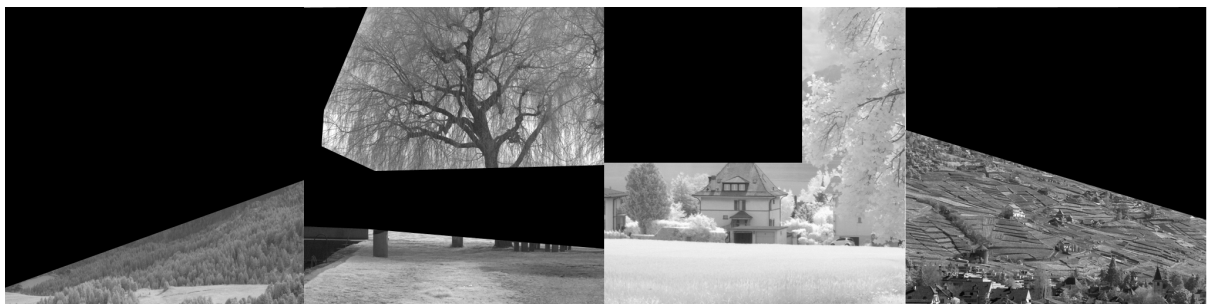

Fig. 11. Masked images to indicate haze regions, which are filled with black colors.

combined with the visible color images using multi-resolution representation, which means that this method did not reflect the haze degradation model. Based on the multi-resolution representation, haze is well removed and fine details are preserved. However, the produced colors are unnatural and diluted, as shown in red boxes of Figs. 7(b), 9(b), and 10(b). This is caused by the discrepancy problem between the near-infrared gray images and visible color images. Please note that near infrared dehazing method was supposed to remove not only the haze, but the color degradation as well. Figures 7(c), 8(c), 9(c), and 10(c) show the dehazed image using another image-pair-based dehazing method [17] where the gradient difference regularization is used to transfer the edges of the near-infrared gray images to the initially dehazed images of the Figs. 7(a), 8(a), 9(a), and 10(a). Note that the initial dehazed images are already refined via alpha matting, which is totally different from the previous section. As expected, the use of the gradient difference regularization can produce sharp edges while preserving the visual color appearances of the initial dehazed images. However, the gradient difference regularization is highly subject to the visual quality level of the initial dehazed image. In other words, the gradient difference regularization can strengthen the edges of the initial dehazed images, however it cannot change the overall color appearances of the initial dehazed images. For example, the edges in the green boxes of the Figs. 7(c) and 9(c) are more strengthened. However, the dehazed images are still so dark, which indicates that the gradient difference regularization cannot change the overall color appearances of initial dehazed images. For these reasons, haze can remain on the dehazed images; see the red boxes of Figs. 9(c) and 10(c). In comparison, the proposed method can provide natural looking colors and fine details. This reaches a conclusion that the proposed color regularization is effective to remove the color distortion and haze at the same time.

*D. Quantitative evaluation*

Three types of measures: image structure similarity (ISS) [22], color difference (CD) [4], and colorfulness (CF) [23] are used to evaluate the near-infrared dehazing methods. The three types of the measures are defined in Table I. As mentioned in Introduction, color distortion and haze degradation are the main issue for near-infrared dehazing. To reflect this point, the three types of measures are adopted. The ISS is used to measure how well the haze is removed on the dehaze images. The CD is used to examine how accurately the dehazed colors are produced. The CF is needed to measure how much the produced colors are diluted. The ISS and CD are the full-reference image quality evaluations, whereas the CF is no reference image quality evaluation. In other words, reference images should be provided to measure the ISS and CD but not necessarily for CF. However, the captured visible color images and near-infrared gray images cannot be used as the reference images. This is because the captured visible color images contain the haze degradation and the captured near-infrared gray images have no colors. To solve this issue, masked images to indicate the haze regions are used in this paper. The examples of the masked images are provided in Fig. 11, where the haze regions are filled with black colors. The haze regions should be manually specified by users. As for the haze regions, the captured near-infrared gray images could be used as the reference images to check how the haze is well removed on dehazed images. As for the non-haze regions, the captured visible color image could be used for the reference images to check how the dehazed colors deviate from the reference images. In other words, for the haze regions, the ISS will be measured using near-infrared gray images (i.e., reference images) and dehazed images (i.e., test images), whereas for non-haze regions, the CD and CF will be measured using visible color images (i.e., reference images) and dehazed images (i.e., test images).

Table II shows the quantitative evaluation for the conventional and proposed methods. In Table II, the ISS value ranges between 0 and 1. Here, '1' indicates the perfect match in the image structures between input images, whereas '0' indicates perfect mismatch. The CD value is greater than or equal to zero. When the colors of input images are identical, the CD value will be zero and it will increase as the colors become more different. The CF will have a small value when an input test image looks grayish. As shown in the last column of Table II, the single dehazing method using dark channel prior [8] has the lowest average ISS value, which means that the single dehazing method fails to remove the haze. As discussed in section of V.A, the dark channel prior has a limitation in that it is difficult to provide good initial points for the unknown haze-free images. In contrast, the image-pair-based dehazing method using multiresolution representation [6] can obtain the highest average ISS value, however its CF values are the lowest. This indicates that the produced colors are relatively diluted. This method just combines the visible color images with the details of the near-infrared gray images based on the multiresolution representation. In this method, the haze degradation model is excluded, and thus it can obtain the highest ISS values. However, this method is not free from the discrepancies in brightness and image structures between the visible color and near-infrared images. Thus, the lowest CF values can be obtained, therefore, the produced colors are unnatural and diluted; See the Figs. 7(b), 8(b), 9(b), and 10(b). Another image-pair-based dehazing method using gradient difference regularization [17] has the highest (or the worst) average CD value, which is caused by the dark colors and enhanced colors in the dehazed images. Compared to the single image dehazing [8], the ISS values can be improved, due to the use of the gradient difference regularization. In contrast to the methods discussed above, both the near-infrared coloring method [12] and the proposed near-infrared dehazing method can obtain good ISS, CD, as well as the CF values. However, the proposed method has smaller average CD value than the near-infrared coloring. This is caused by different goals of the near-infrared coloring and the proposed near-infrared dehazing. As discussed in the section of V.B, the near-infrared coloring adds colors to near-infrared gray images, and thus the visual appearances of the colored near-infrared images are influenced by the overall brightness of the captured near-infrared gray images. On the other hand, the proposed method includes the haze degradation model, thus the visual color appearances of the dehazed images are derived from the captured visible color images. As result, the proposed method can have smaller CD values than the near-infrared coloring. This is the main difference between the near-infrared coloring and dehazing. Also, the sky, cloud, and shadow descriptions of the proposed method are better than those of the near-infrared coloring, which cannot be evaluated by the three measures. The proposed method has the smallest average CD value and obtains high ISS values. This indicates that the proposed color regularization is effective to remove the haze and color distortion problems. The conventional and proposed dehazing methods have their own merits and demerits. However, the conventional dehazing methods except the near-infrared coloring have the worst case for one of the ISS, CD, and CF, which would significantly deteriorate the overall visual quality. In contrast, the proposed method can obtain good ISS, CD, and CF values. In other words, the proposed method can remove haze and preserve fine detail while avoiding color degradation. This is the advantage of the proposed method.

Note that the CF values of the single image dehazing method [8] are higher than the proposed method. However, the CD values of the proposed method are smaller than those of the single image dehazing method. The CD measure includes the chroma difference between tested images and it is well-known that the chroma is highly related to the colorfulness [4]. In Table I, the CF measure includes the chroma. This means that the CF values of the proposed method would be closer to the reference CF values than those of the single image dehazing method. In other words, we cannot say that higher CF values correspond to better image quality. However, it is expected that the reference CF values could be closer to the CF values of the proposed method.

## VII. CONCLUSIONS

In this paper, a new near-infrared dehazing model based on color regularization is proposed. The discrepancies in brightness and image structures between captured near-infrared and visible color images generate another color distortion in dehazed images. Therefore, the color distortion and haze degradation should be considered at the same time. It motivates us to develop the new color regularization method, which can model the unknown colors of the haze-free images in the haze degradation model. Experimental results showed that the proposed color regularization can solve the color distortion and haze degradation problems at the same time. Also, it is shown that the proposed color regularization can provide better visual appearance in color and details than the conventional regularizations: dark channel, and gradient and gradient difference regularizations. Furthermore, visual effects according to the use of the near-infrared dehazing and coloring are compared to help understand the differences between the near-infrared dehazing and near-infrared coloring.